\documentclass[sigconf,natbib=false]{acmart}
\AtBeginDocument{%
  }

\usepackage{booktabs,multirow,makecell}
\usepackage[table]{xcolor}
\usepackage{xfp,xcolor}
\usepackage{siunitx}
\usepackage{threeparttable}
\usepackage[ruled,vlined]{algorithm2e}
\usepackage{hhline}
\usepackage{placeins}
\usepackage{float}
\usepackage{enumitem}
\usepackage{nicematrix}
\newcommand{\pctinc}[2]{\fpeval{round(((#1)-(#2))/(#2)*100,1)}}
\newcommand{\gain}[2]{\textbf{\textcolor{black}{(↑\pctinc{#1}{#2}\%)}}}
\definecolor{rowgray}{gray}{0.92}

\copyrightyear{2026}
\acmYear{2026}
\setcopyright{cc}
\setcctype{by}
\acmConference[SAC '26]{The 41st ACM/SIGAPP Symposium on Applied Computing}{March 23--27, 2026}{Thessaloniki, Greece}
\acmBooktitle{The 41st ACM/SIGAPP Symposium on Applied Computing (SAC '26), March 23--27, 2026, Thessaloniki, Greece}
\acmPrice{}
\acmDOI{10.1145/3748522.3779742}
\acmISBN{979-8-4007-2294-3/2026/03}

\RequirePackage[
  datamodel=acmdatamodel,
  style=acmnumeric,
  ]{biblatex}

\begin{document}

\title{Memory Bank Compression for \\Continual Adaptation of Large Language Models}
% Please make sure that the short title does not exceed the width of one column
\renewcommand{\shorttitle}{Memory Bank Compression for Continual Adaptation of LLMs}

\author{Thomas Katraouras}
 \email{tkatraouras@uth.gr}
 \affiliation{%
   \institution{University of Thessaly}
   \city{Volos}
   \country{Greece}
}

\author{Dimitrios Rafailidis}
 \email{draf@uth.gr}
 \affiliation{%
   \institution{University of Thessaly}
   \city{Volos}
   \country{Greece}
}

%This command displays author info in page headers
% Please use the following convention:
% One author: J. Smith
% Two authors: J. Smith and I. Jones
% Three and more authors: J. Smith et al.
\renewcommand{\shortauthors}{T. Katraouras and D. Rafailidis}

\begin{abstract}
Large Language Models (LLMs) have become a mainstay for many everyday applications. However, as data evolve their knowledge quickly becomes outdated. Continual learning aims to update LLMs with new information without erasing previously acquired knowledge. Although methods such as full fine-tuning can incorporate new data, they are computationally expensive and prone to catastrophic forgetting, where prior knowledge is overwritten. Memory-augmented approaches address this by equipping LLMs with a memory bank, that is an external memory module which stores information for future use. However, these methods face a critical limitation, in particular, the memory bank constantly grows in the real-world scenario when large-scale data streams arrive. In this paper, we propose MBC, a model that compresses the memory bank through a codebook optimization strategy during online adaptation learning. To ensure stable learning, we also introduce an online resetting mechanism that prevents codebook collapse. In addition, we employ Key-Value Low-Rank Adaptation in the attention layers of the LLM, enabling efficient utilization of the compressed memory representations. Experiments with benchmark question–answering datasets demonstrate that MBC reduces the memory bank size to 0.3\% when compared against the most competitive baseline, while maintaining high retention accuracy during online adaptation learning. Our code is publicly available at https://github.com/Thomkat/MBC.
\end{abstract}




\maketitle

\section{Introduction}\label{sec:Introduction}
Large Language Models (LLMs)~\cite{openaiGPT4TechnicalReport2024,touvronLlama2Open2023a} have shown strong performance on a wide range of natural language processing tasks, including machine translation~\cite{wangmachintranslation2024}, summarization~\cite{vanveensummarization2024}, question answering~\cite{singhalquestionanswering2025}, and advanced reasoning~\cite{yaoreasoning2023}. They are now widely employed in many applications such as search engines~\cite{zhusearchengines2024} and personal assistants~\cite{gunterAppleIntelligenceFoundation2024}. However, a major limitation of these models is that they are static~\cite{jangstaticmodels2022}. Once trained, their parameters reflect only the data seen during training, and they cannot easily incorporate new knowledge. This leads to the problem of knowledge cutoff, where the model’s internal knowledge becomes outdated as new information appears~\cite{openaiGPT4TechnicalReport2024,touvronLlama2Open2023a}.

To address this limitation, Retrieval-Based Augmentation (RAG) strategies have been introduced~\cite{lewisRAG2020,wuRAGsurvey2024}. A frozen LLM uses a retriever to fetch relevant passages from an external corpus at inference time, providing the model with access to up-to-date information without retraining~\cite{lewisRAG2020}. However, RAG methods face several challenges. They depend on nearest-neighbor search, which adds computational overhead and latency~\cite{wuRAGsurvey2024}. The quality of the retrieval affects the LLM performance, and errors in retrieval propagate directly to the generator~\cite{zhaoRAGquality2025}. Retrievers also often require domain-specific tuning and may struggle to generalize across domains~\cite{misrahiragoutofdomain2025}. Furthermore, the retrieved passages are typically concatenated with the query in the input context window, limiting the model’s ability to fully utilize the information and creating issues when the combined length exceeds the model’s capacity~\cite{lewisRAG2020}. These issues limit the scalability of RAG for long-term adaptation in streaming environments, reflecting on the real-world scenario.

To solve the problem of LLMs' long-term adaptation, continual learning methods have been proposed~\cite{continual_2,continual_1}. In this setting, models are updated as new data arrive. The simplest approach, full fine-tuning, also known as uniform fine-tuning~\cite{huCamels} since all tokens are weighted equally, updates all parameters on the new data. While effective for small models, this is computationally expensive for large LLMs and is prone to catastrophic forgetting~\cite{catastrophic_forgetting}, where performance on previously learned knowledge degrades as the model is optimized on new information. Parameter-efficient fine-tuning (PEFT) methods~\cite{xupeftMethods2023} such as adapters~\cite{houlsbyadapters12019,pfeifferadapters22021}, prefix-tuning~\cite{prefix_tuning}, and LoRA~\cite{huLoRALowRankAdaptation2021} address the computational cost by introducing small trainable modules while keeping most parameters frozen. These approaches reduce the training overhead, however, they still require gradient-based updates at deployment time, which is impractical in streaming scenarios. Additional strategies have been proposed to make updates selective and stable, for example by restricting updates to predefined salient spans~\cite{salientspansbaseline}, by meta-learning token importance weights~\cite{huCamels}, or by interleaving past and new examples through replay~\cite{chaudhryreplay12019,schwarzreplay22018}. Despite these refinements, the fundamental limitations persist, namely the need for repeated optimization, substantial computational and latency overhead, and continued vulnerability to catastrophic forgetting~\cite{continual_2}.

A promising direction for continual learning of LLMs is memory augmentation~\cite{memory_1,memory_2,memory_3}. Instead of retrieving raw text from an external corpus, information is stored directly in a structured memory module, which can also be updated dynamically. At inference time, the model can draw on these stored representations to adapt its behavior and incorporate new information. This avoids repeated gradient updates and provides a direct connection between the stored knowledge and model’s computations~\cite{memory_1}. However, memory augmentation introduces challenges as more documents are processed, the memory bank constantly grows. This increases storage costs and slows down inference, since the model must attend to an ever-growing set of contexts~\cite{memory_4}. Expanding memory without disrupting the model’s original behavior remains difficult, and as a consequence, the memory augmentation strategies require retraining or fine-tuning to remain effective.

More recently, to overcome this shortcoming, memory-augmented frameworks have been proposed that store learned modulation parameters for each document in an external memory bank~\cite{tack2024mac}. These approaches keep the base model frozen, condition the model on the entire memory bank rather than a single retrieved document, and avoid further fine-tuning during adaptation, while mitigating catastrophic forgetting. Nevertheless, in the real-world scenario where the document stream reaches hundreds of thousands or millions of entries, the memory bank grows very large and becomes difficult to manage. This highlights scalability as an open problem in memory-augmented systems, alongside the need to balance adaptation, efficiency, and stability.

In this paper, we propose MBC, a model that compresses the memory bank while maintaining high performance on downstream question-and-answer (QA) tasks. Specifically, we make the following contributions:

\begin{itemize}[topsep=1pt,itemsep=1pt]
    \item We propose a memory bank compression method based on a codebook optimization strategy, which stores indices to this codebook instead of full document representations. In addition, we introduce an online resetting strategy to prevent codebook collapse and ensure balanced code utilization and stable training.
    \item We employ Key-Value Low-Rank Adaptation targeted only to the attention layers of the model. In doing so, we improve the proposed model's ability to adapt when new data arrive without requiring full fine-tuning.
\end{itemize}

We conduct experiments on benchmark QA datasets, comparing our MBC model with baseline methods. Our results demonstrate that MBC significantly reduces the memory bank size to 0.3\%, when compared with the initial size of baseline strategies, and improves the QA accuracy. Furthermore, our model maintains a high retention accuracy when evaluated for catastrophic forgetting during the challenging online adaptation scenario.

The remainder of the paper is structured as follows: Section \ref{sec:problem_form} formulates the problem of online learning in LLMs. Section \ref{sec:Proposed Model} outlines the MBC model and Section \ref{sec:Experimental Evaluation} provides the experimental evaluation. Finally, Section \ref{sec:Conclusion} concludes the paper, summarizing key findings and discussing potential future directions.

\section{Online Adaptation of LLMs}
\label{sec:problem_form}
Let $f_{\theta_b}$ be a pretrained and outdated language model with parameters $\theta_b$. During online adaptation, $f_{\theta_b}$ is continuously updated using a stream of new documents, denoted as $D^{test} := \{ d_i \}$. The adaptation process yields an updated model $f_{\tilde{\theta}_b}$~\cite{huCamels}. This adapted model is then evaluated on a set of queries $Q^{test} := \{ q_i \}$ paired with labels $Y^{test} := \{ y_i \}$. Each query-label pair $(q_i, y_i)$ is assumed to be sampled from a distribution conditioned on the corresponding document $d_i$, i.e., $(q_i, y_i) \sim p(q, y \mid d_i)$. For example, in a QA setting, $q_i$ may represent a question about information contained in $d_i$, with $y_i$ being the correct answer~\cite{tack2024mac}. During adaptation with $D^{test}$, the related queries $Q^{test}$ remain inaccessible to the model. Therefore, the update procedure must be query-agnostic. To this end, we assume access to an auxiliary training set $D^{train}$ with associated queries $Q^{train}$ and labels $Y^{train}$, defined in the same way as $(Q^{test}, Y^{test})$. This auxiliary set provides examples of query–document relationships and guides the model in updating its parameters while retaining past knowledge and improving on future queries~\cite{huCamels}. The training step involving $(D^{train}, Q^{train}, Y^{train})$ is the learning phase. The subsequent process of updating $f_{\theta_b}$ using the test stream $D^{test}$, without access to queries, is the online adaptation phase~\cite{tack2024mac}.

\section{Proposed Model}\label{sec:Proposed Model}
\subsection{Memory of Aggregated Contexts}
The proposed model has three core components: (i) an amortization network that encodes documents, (ii) a memory bank that stores the encoded information, and (iii) an aggregation network that synthesizes the stored information to answer a given query.

\paragraph{Amortization Network} The amortization network is responsible for mapping each document into a compact latent representation that can be efficiently stored and retrieved~\cite{tack2024mac}. Formally, this network is denoted $g_{\theta_{amort}}$, implemented with a T5 encoder-decoder model~\cite{t5}. Given a document $d_i$, the amortization network produces a continuous latent representation \(\phi_i := g_{\theta_{\text{amort}}}(d_i) \in \mathbb{R}^{T \times D}\), where $T$ is the number of tokens in the representation and $D$ is the hidden dimension of the base model.

\paragraph{Memory Bank} The context vectors $\phi_i$ generated from the document stream are stored in an external memory bank \(\mathcal{M} := \{ \phi_i \mid d_i \in D^{test}\}\)~\cite{tack2024mac}. This bank serves as a growing knowledge base for the base LLM.

\paragraph{Aggregation Network} When a query $q_i$ is presented at test time, the model must retrieve relevant information from the memory bank. An aggregation network, denoted $h_{\psi}$, is trained to perform this function dynamically. It takes the entire memory bank $\mathcal{M}$ and an encoded representation of the current query $g_{\theta_{input}}(q_i)$ as input, where $\theta_{input}$ uses the same architecture as $\theta_{amort}$. The network is permutation-invariant with respect to the ordering of $\mathcal{M}$. Using a cross-attention mechanism~\cite{crossattn,attentionisallyouneed}, it synthesizes the stored context vectors into a single, query-specific modulation \(\phi_i^* := h_{\psi (g_{\theta_{input}}(q_i), \mathcal{M})}\).
This modulation $\phi_i^*$ acts as a set of soft prompts, injected as learnable prefixes into the key–value matrices of each self-attention layer of the base LLM via P-tuning v2~\cite{ptuningv2}. Formally, the modulated base LLM can be expressed as \(f_{\theta_b}^{\phi_i^*}(q_i) := f_{\theta_b}(q_i; \{K_\ell', V_\ell'\}_{\ell=1}^L)\), with $K_\ell', V_\ell'$ denoting the modified key and value matrices in layer $\ell$ after prefixing with $\phi_i^*$. 
To efficiently handle large memory banks at inference time, a hierarchical modulation aggregation strategy is used. In particular, the context set is first partitioned into smaller subgroups, each aggregated individually, and the resulting representations are recursively combined until a single final modulation is obtained. This divide-and-conquer procedure reduces the memory complexity to \(\mathcal{O}(MT)\), where $M$ is a hyperparameter and $T$ the number of tokens, ensuring scalability even as the number of stored documents increases~\cite{tack2024mac}.

\subsection{Codebook Optimization for Memory Bank Compression}  
A critical limitation of storing continuous context vectors $\{\phi_i\}$ is that the memory bank $\mathcal{M}$ can grow very large as the document stream increases. To address this, a Vector Quantised-Variational AutoEncoder (VQ-VAE)–style~\cite{vqvae} quantization module is introduced to compress the memory. Instead of storing the high-dimensional continuous vector $\phi_i$, each context vector is mapped to its nearest entry in a learned finite codebook $E \in \mathbb{R}^{N_c \times D}$, where $N_c$ denotes the number of code vectors: \(c_i := \underset{j}{\operatorname{argmin}} \, \|\phi_i - E_j\|_2^2\). The selected codebook vector is denoted by \(\hat{\phi}_i^{\mathrm{hard}} := E_{c_i}\), and only the integer index $c_i$ is stored, resulting in a compressed memory bank representation \(\mathcal{M}_{\mathrm{VQ}} := \{c_i\}\). As $\hat{\phi}_i^{\mathrm{hard}}$ is discrete and non-differentiable, during the forward pass a straight-through estimator (STE)~\cite{straightthroughestimator} is used to define \(\hat{\phi}_i := \phi_i + \mathrm{sg}[\hat{\phi}_i^{\mathrm{hard}} - \phi_i]\), where $\mathrm{sg}[\cdot]$ denotes the stop-gradient operator. Thus, $\hat{\phi}_i$ takes the value of $\hat{\phi}_i^{\mathrm{hard}}$, still allowing gradients to flow back into $\phi_i$. In subsequent notations, $\hat{\phi}_i$ denotes the differentiable forward-pass representation and $\hat{\phi}_i^{\mathrm{hard}}$ is used only in the vector quantization loss.
The effective size of the memory bank is therefore reduced to the set of indices together with the codebook $E$. The codebook itself is optimized during end-to-end training. During inference, the stored indices are used to retrieve their corresponding quantized vectors $\{\hat{\phi}_i\}$, which are then aggregated by $h_{\psi}$ together with the query representation to produce the modulation $\hat{\phi}_i^*$.

\subsection{Online Codebook Resetting}
\label{sec:codebook_reset_intro}
To prevent underutilization and codebook collapse, where only a small subset of codes is repeatedly used, the codebook is updated during training using an exponential moving average (EMA) of code usage. For a mini-batch of size $K$, let
\begin{equation}
    n_j := \sum_{i=1}^K \mathbf{1}[c_i=j],
    \quad
    u_j := \gamma\,u_j + (1-\gamma)\,n_j \quad \forall j \in \{1,\dots,N_c\}
\label{eq:resetting}
\end{equation}
where $u_j$ tracks smoothed usage and $\gamma\!\in(0,1)$ is a decay rate hyperparameter. Codes with usage below a hyperparameter threshold $\epsilon$ are marked as inactive: \(I_{\mathrm{dead}} := \{\, j \mid u_j < \epsilon \,\}\). When $I_{\mathrm{dead}}\neq\emptyset$, up to $|I_{\mathrm{dead}}|$ distinct encoder outputs are sampled from the current batch, $\{\phi_{s(j)}\}$, to reinitialize the corresponding codebook vectors:
\begin{equation}
    E_j := \phi_{s(j)} \quad \forall j \in I_{\mathrm{dead}},
    \qquad
    u_j := \bar u := \frac{1}{N_c}\sum_{\ell=1}^{N_c} u_\ell
\end{equation}
where $s(j)$ denotes indices sampled uniformly and randomly without replacement from the batch, and $\bar u$ denotes the mean usage across all codes, ensuring that reinitialized entries retain a non-negligible prior usage estimate. This procedure, applied only during training (without gradients), maintains codebook diversity and prevents collapse, as we will experimentally show in Section~\ref{sec:codebook_reset}.

\subsection{Key-Value Low-Rank Adaptation for Modulation Adaptation}  
The modulation $\hat{\phi}_i^*$ is injected into the key–value pairs of each self-attention layer of the base LLM. Instead of keeping the base LLM entirely frozen such as the study reported in~\cite{tack2024mac}, we introduce lightweight Low-Rank Adaptation (LoRA)~\cite{huLoRALowRankAdaptation2021} modules specifically into the key and value projections (KV-LoRA).
Here, $\theta_b$ denotes the frozen parameters of the pretrained model $f_{\theta_b}$. 
A small set of trainable parameters $\theta_{\mathrm{KV\text{-}LoRA}}$ is added, parameterizing low-rank updates to the modulated key and value matrices:
\begin{equation}
    K_\ell'' := K_\ell' + A_{K,\ell} B_{K,\ell}, 
    \qquad 
    V_\ell'' := V_\ell' + A_{V,\ell} B_{V,\ell}
\end{equation}
where $A_{K,\ell}, A_{V,\ell} \in \mathbb{R}^{D \times r}$ and 
$B_{K,\ell}, B_{V,\ell} \in \mathbb{R}^{r \times D}$ are low-rank factors 
with rank $r \ll D$.
In practice, the updates are scaled by a factor $\alpha / r$ and regularized with dropout. 
KV-LoRA is applied only to the final $n_{lora}$ transformer layers, balancing computational efficiency with adaptation capacity.
$r$, $\alpha$, $n_{lora}$, and the dropout probability $\rho$ are hyperparameters.
The adapted model thus has parameters $\tilde{\theta}_b := \theta_b \cup \theta_{\mathrm{KV\text{-}LoRA}}$, which preserves pretrained knowledge allowing the attention mechanism to more effectively exploit the modulation $\hat{\phi}_i^*$.  
Formally, the modulated base LLM is expressed as:
\begin{equation}
    f_{\tilde{\theta}_b}^{\hat{\phi}_i^*}(q_i) := f_{\tilde{\theta}_b}(q_i; \{{K''}_\ell, {V''}_\ell\}_{\ell=1}^L)
\end{equation}

\subsection{End-to-End Training Objective}  
The entire architecture is trained end-to-end, including the amortization and aggregation networks, the VQ codebook, and the KV-LoRA modules. 
The primary objective is a question-answering (QA) loss $\mathcal{L}_{\mathrm{QA}}$, defined as the negative log-likelihood of predicting the target sequence $y_i$ conditioned on the query $q_i$, the corresponding document $d_i$, and its modulation $\phi_i^*$. Answer generation is carried out by the base LLM $f_{\tilde{\theta}_b}^{\hat{\phi}_i^*}(q_i)$: \(\mathcal{L}_{\mathrm{QA}} := - \log p_{\tilde{\theta}_b}^{\hat{\phi}_i^*}(y_i \mid q_i, d_i)\). To enforce a compact and discrete memory representation, a vector quantization loss $\mathcal{L}_{\mathrm{VQ}}$ is used. 
Given a continuous context vector $\phi_i \in \mathbb{R}^{T \times D}$, its nearest codebook entry is denoted by $\hat{\phi}_i^{\mathrm{hard}} := E_{c_i}$.   
The quantization loss is defined using $\hat{\phi}_i^{\mathrm{hard}}$ to ensure proper updates of both the codebook and encoder:
\begin{equation}
    \mathcal{L}_{\mathrm{VQ}} 
    := \| \mathrm{sg}[\phi_i] - \hat{\phi}_i^{\mathrm{hard}} \|_2^2 
       + \beta \, \| \phi_i - \mathrm{sg}[\hat{\phi}_i^{\mathrm{hard}}] \|_2^2
\end{equation}
where $\beta > 0$ is the commitment cost hyperparameter. 
The first term updates the selected codebook entries $E_{c_i}$, while the second one encourages encoder outputs $\phi_i$ to remain close to their quantized assignments.
The final objective is a weighted combination: \(\mathcal{L}_{\mathrm{total}} = \mathcal{L}_{\mathrm{QA}} + \lambda_{\mathrm{VQ}} \, \mathcal{L}_{\mathrm{VQ}}\), where $\lambda_{\mathrm{VQ}}$ is a hyperparameter that balances the influence of the quantization loss.
During training, the parameters of the amortization network ($\theta_{\mathrm{amort}}$), input encoder ($\theta_{\mathrm{input}}$), aggregation network ($\psi$), KV-LoRA modules ($\theta_{\mathrm{KV\text{-}LoRA}}$), and the codebook ($E$) are optimized end-to-end, while the base model parameters $\theta_b$ remain frozen: 
\begin{equation}
    \min_{\theta_{\mathrm{amort}}, \theta_{\mathrm{input}}, \psi, \theta_{\mathrm{KV\text{-}LoRA}}, E} 
    \frac{1}{K} \sum_{i=1}^K 
    \Big[ \mathcal{L}_{\mathrm{QA}}(q_i, d_i, y_i) + \lambda_{\mathrm{VQ}} \, \mathcal{L}_{\mathrm{VQ}}(\phi_i, \hat{\phi}_i^{\mathrm{hard}}) \Big]
\end{equation}
where K is the batch size. Optimization is performed using the Adam~\cite{adam} optimizer. An overview of the MBC end-to-end optimization algorithm is presented in Algorithm~\ref{alg:training_detailed}.

\subsection{Online Adaptation of MBC}  
After training is completed, the online adaptation phase follows.
This phase requires no gradient-based updates and operates entirely through forward passes. 
The procedure consists of two components:

\paragraph{Memorization.} 
For each new document $d_i$ arriving in the test stream $D^{\mathrm{test}}$, the amortization network $g_{\theta_{\mathrm{amort}}}$ encodes $d_i$ into a context vector $\phi_i$, which is subsequently quantized to the nearest codebook entry in $E$. 
The resulting discrete code $c_i$ is stored in the compressed memory bank $\mathcal{M}_{\mathrm{VQ}}$.

\paragraph{Inference.} 
When a query $q_i$ is received, the model uses the stored codes $\{c_j\}$ from the memory bank and retrieves the corresponding quantized context vectors $\{\hat{\phi}_j\}$ from the codebook $E$. 
These vectors are aggregated by $h_{\psi}$ together with the query representation $g_{\theta_{\mathrm{input}}}(q_i)$ to produce a query-specific modulation $\hat{\phi}_i^*$. 
This modulation conditions the KV-LoRA-augmented base LLM $f_{\tilde{\theta}_b}(q_i)$, which then generates the final answer.
The online adaptation and evaluation procedure is presented in Algorithm~\ref{alg:online_adaptation}.

\begin{algorithm}[!htbp]
\footnotesize
\caption{MBC End-to-End Optimization}
\label{alg:training_detailed}
\LinesNumbered
\KwIn{Amortization params $\theta_{\mathrm{amort}}$, input encoder params $\theta_{\mathrm{input}}$, base LLM params $\theta_b$, aggregation params $\psi$, KV-LoRA params $\theta_{\mathrm{KV\text{-}LoRA}}$, hidden dimension $D$, training corpus $D^{\mathrm{train}}$, learning rate $\eta$, epochs $m$, batch size $K$, VQ commitment cost $\beta_{\text{commit}}$, VQ weight $\lambda_{\mathrm{VQ}}$, codebook size $N_c$, reset threshold $\epsilon$, EMA decay rate $\gamma$}
\KwOut{$\theta_{\mathrm{amort}}, \theta_{\mathrm{input}}, \psi, \theta_{\mathrm{KV\text{-}LoRA}}, E$}

\tcp{Initialize codebook and usage EMA}
$E_{j} \sim \mathcal{U}\!\big(-\tfrac{1}{N_c}, \tfrac{1}{N_c}\big)\ \ \forall j \in \{1,\dots,N_c\}$,\quad $u_j \gets 0\ \ \forall j$

\For{epoch $= 1 \to m$}{
  Sample documents $\{d_1,\dots,d_K\} \subset D^{\mathrm{train}}$

    Sample QA pairs $(q_i, y_i) \sim p(q,y \mid d_i)$ for $i=1,\dots,K$
    
  \For{$i = 1$ $\to$ $K$}{
    $\phi_i \gets g_{\theta_{\mathrm{amort}}}(d_i)$ \tcp*{Encode document}

    \tcp{Vector quantization (nearest code)}
    $c_i \gets \arg\min_{j \in \{1,\dots,N_c\}} \|\phi_i - E_j\|_2^2$, \quad $\hat{\phi}_i^{\mathrm{hard}} \gets E_{c_i}$

    $\hat{\phi}_i \gets \phi_i + \mathrm{sg}[\hat{\phi}_i^{\mathrm{hard}} - \phi_i]$ \tcp*{Straight-through estimator}
  }

  \tcp{Update code usage EMA and reset dead codes}
  $n_j \gets \sum_{i=1}^{K} \mathbf{1}[c_i = j]\ \ \forall j$,\quad 
    $u_j \gets \gamma\,u_j + (1-\gamma)\,n_j\ \ \forall j$
    
    $I_{\mathrm{dead}} \gets \{\, j \in \{1,\dots,N_c\} \mid u_j < \epsilon \,\}$

  \If{$|I_{\mathrm{dead}}| > 0$}{
    \tcp{Replace dead codes with random batch samples}
    $S \sim Unif \!\Big(\tbinom{\{1,\dots,K\}}{|S|}\Big), 
\quad |S|=\min(|I_{\mathrm{dead}}|,K)$

$E_j \gets \phi_{s(j)} 
\quad \forall j \in I_{\mathrm{dead}},\ \text{up to } |S|$

      $u_j \gets \tfrac{1}{N_c}\sum_{\ell=1}^{N_c} u_\ell 
      \quad \forall j \in I_{\mathrm{dead}}$ \tcp*{Reset usage}
  }

  \tcp{Aggregate quantized contexts with the query}
  $\hat{\phi}_i^{*} \gets h_{\psi}\!\big(g_{\theta_{\mathrm{input}}}(q_i), \{\hat{\phi}_i\}_{i=1}^{K}\big)$

  \tcp{QA loss via modulated base LLM}
  $\mathcal{L}_{\mathrm{QA}} \gets \frac{1}{K}\sum_{i=1}^{K} \mathrm{CrossEntropy}\!\big(f_{\tilde{\theta}_b}^{\hat{\phi}_i^{*}}(q_i),\, y_i\big)$

  \tcp{VQ loss (codebook and commitment terms)}
  $\mathcal{L}_{\text{codebook}} \gets \frac{1}{K}\sum_{i=1}^{K} \big\|\mathrm{sg}[\phi_i] - \hat{\phi}_i^{\mathrm{hard}}\big\|_2^2$

  $\mathcal{L}_{\text{commit}} \gets \frac{1}{K}\sum_{i=1}^{K} \big\|\phi_i - \mathrm{sg}[\hat{\phi}_i^{\mathrm{hard}}]\big\|_2^2$

  $\mathcal{L}_{\mathrm{VQ}} \gets \mathcal{L}_{\text{codebook}} + \beta_{\text{commit}}\cdot \mathcal{L}_{\text{commit}}$

  $\mathcal{L}_{\mathrm{total}} \gets \mathcal{L}_{\mathrm{QA}} + \lambda_{\mathrm{VQ}}\, \mathcal{L}_{\mathrm{VQ}}$ \tcp*{Total objective}

  \tcp{Gradient updates (base $\theta_b$ frozen)}
  $\theta_{\mathrm{amort}} \gets \theta_{\mathrm{amort}} - \eta \nabla_{\theta_{\mathrm{amort}}} \mathcal{L}_{\mathrm{total}}$

  $\theta_{\mathrm{input}} \gets \theta_{\mathrm{input}} - \eta \nabla_{\theta_{\mathrm{input}}} \mathcal{L}_{\mathrm{total}}$

  $\psi \gets \psi - \eta \nabla_{\psi} \mathcal{L}_{\mathrm{total}}$

  $\theta_{\mathrm{KV\text{-}LoRA}} \gets \theta_{\mathrm{KV\text{-}LoRA}} - \eta \nabla_{\theta_{\mathrm{KV\text{-}LoRA}}} \mathcal{L}_{\mathrm{total}}$

  $E \gets E - \eta \nabla_{E} \mathcal{L}_{\mathrm{total}}$
}
\end{algorithm}

\begin{algorithm}[!htbp]
\footnotesize
\caption{Online Adaptation of MBC}
\label{alg:online_adaptation}
\LinesNumbered
\KwIn{Test document stream $D^{\mathrm{test}}$, test QA set $\{(q_i, y_i)\}_{i=1}^I$, amortization params $\theta_{\mathrm{amort}}$, input encoder params $\theta_{\mathrm{input}}$, base LLM params with KV-LoRA $\tilde\theta_b$, aggregation params $\psi$, learned codebook $E$}
\KwOut{EM and F1 over $\{(q_i,y_i)\}_{i=1}^I$}

$\mathcal{M}_{\mathrm{VQ}} \gets \emptyset$ \tcp*{Initialize compressed memory bank}

\For{$d_k \in D^{\mathrm{test}}$}{
  $\phi_k \gets g_{\theta_{\mathrm{amort}}}(d_k)$ \tcp*{Encode document}

  $c_k \gets \arg\min_{j} \|\phi_k - E_j\|_2^2$ \tcp*{Quantize}

  $\mathcal{M}_{\mathrm{VQ}} \gets \mathcal{M}_{\mathrm{VQ}} \cup \{c_k\}$ \tcp*{Save document to bank}
}

\For{$i = 1 \to I$}{
  $\hat{\phi}_i^* \gets h_{\psi}\!\big(g_{\theta_{\mathrm{input}}}(q_i), \{E_{c_j}\}_{c_j \in \mathcal{M}_{\mathrm{VQ}}}\big)$ \tcp*{Aggregate memory with query}
  
  $\hat{y}_i \gets f_{\tilde{\theta}_b}^{\hat{\phi}_i^*}(q_i)$ \tcp*{Predict answer}
}

\tcp{Final evaluation; norm(): lowercase, strip punctuation, remove articles, collapse whitespace}
$\text{EM} \gets \frac{1}{I} \sum_{i=1}^{I} \mathbf{1}\!\big[\mathrm{norm}(y_i) = \mathrm{norm}(\hat{y}_i)\big]$ \tcp*{Exact Match}

$\text{F1} \gets \frac{1}{I} \sum_{i=1}^{I} \mathrm{F1}_{\text{token}}(y_i,\hat{y}_i)$ \tcp*{Token-Level F1}

\Return $(\text{EM}, \text{F1})$
\end{algorithm}

\section{Experimental Evaluation}\label{sec:Experimental Evaluation}
\subsection{Datasets}
Following~\cite{huCamels,tack2024mac}, we evaluate the examined models on three QA datasets:
\paragraph{StreamingQA~\cite{StreamingQA_dataset}} It contains questions created by annotators or generated with language models. Questions are based on timestamped English WMT news articles (2007–2020), which are also included in the dataset. Following prior setups, we use 21K training, 1.7K validation, and 5K test questions, along with the same number of documents. For QA pre-training baselines, 40K training and 4K validation questions are used.
\paragraph{SQuAD~\cite{Squad_dataset}} The Stanford Question Answering Dataset (SQuAD) includes crowdsourced questions on Wikipedia, where answers are spans within the article. Following prior setups, we use 39.9K training, 5.6K validation, and 10.6K test questions, with 8.6K training, 1.2K validation, and 2.1K test documents, respectively. For QA pre-training baselines, 40K training and 2.1K validation questions are used.
\paragraph{ArchivalQA~\cite{ArchivalQA_dataset}} It is built from New York Times Annotated Corpus articles~\cite{nytcorpus} with questions generated using language models. Answers are text spans within the articles. Following prior setups, we use 21.7K training, 5.3K validation, and 8.7K test questions, with 12.8K training, 3.0K validation, and 5.0K test documents, respectively. For QA pre-training baselines, 12.4K training and 3K validation questions are used.

\subsection{Evaluation Protocol}  
We follow the training configuration of prior works~\cite{huCamels,tack2024mac} for fair comparison across baselines. 
For each dataset, the model is adapted using 1,665 documents sampled from the test stream $D^{\mathrm{test}}$, after which its performance is evaluated on QA pairs drawn from the same documents.
We report Exact Match (EM) and token-level F1 scores as evaluation metrics.

The EM score measures the fraction of predictions that exactly match the ground-truth answer after normalization (lowercasing, punctuation and article removal, collapsing multiple spaces into one):
\begin{equation}
    \mathrm{EM} = \frac{1}{I}\sum_{i=1}^I \mathbf{1}\!\big[\mathrm{norm}(\hat{y}_i) = \mathrm{norm}(y_i)\big]
\end{equation}
where $I$ is the number of QA pairs, $\hat{y}_i$ is the predicted answer, and $y_i$ is the ground truth.

The token-level F1 score measures the harmonic mean of precision and recall at the token level:
\begin{equation}
    \mathrm{Precision}_i = \frac{|\,\mathrm{tok}(\hat{y}_i) \cap \mathrm{tok}(y_i)\,|}{|\mathrm{tok}(\hat{y}_i)|},
    \qquad
    \mathrm{Recall}_i = \frac{|\,\mathrm{tok}(\hat{y}_i) \cap \mathrm{tok}(y_i)\,|}{|\mathrm{tok}(y_i)|}
\end{equation}
\begin{equation}
    \mathrm{F1}_i = \frac{2 \cdot \mathrm{Precision}_i \cdot \mathrm{Recall}_i}{\mathrm{Precision}_i + \mathrm{Recall}_i},
    \qquad
    \mathrm{F1} = \frac{1}{I}\sum_{i=1}^I \mathrm{F1}_i
\end{equation}
where $\mathrm{tok}(\cdot)$ denotes the tokenized representation of the answer.

\subsection{Experimental Setup}
\subsubsection{Implementation Details}  
We evaluate MBC using four backbone LLMs: the GPT-2 family (DistilGPT2~\cite{distilgpt2_paper}, GPT2-Large~\cite{gpt2_paper}, GPT2-XL~\cite{gpt2_paper}) and LLaMA-2-7B~\cite{touvronLlama2Open2023a}, with 82M, 774M, 1.5B, and 7B parameters, respectively. The amortization network $g_{\theta_{\mathrm{amort}}}$ is based on T5~\cite{t5}, using T5-Small for DistilGPT2, T5-Base for GPT2-Large, and T5-Large for GPT2-XL and LLaMA-2-7B. The input encoder $g_{\theta_{\mathrm{input}}}$ uses T5-Small for DistilGPT2 and T5-Base for the rest. The amortization network outputs $T=12$ tokens for DistilGPT2 and 24 for the rest. The aggregation network $h_{\psi}$ consists of four cross-attention blocks~\cite{crossattn,attentionisallyouneed}, where $g_{\theta_{\mathrm{input}}}(q_i)$ provides the initial query, the memory bank $\mathcal{M}_{\mathrm{VQ}}$ provides keys and values, and subsequent blocks take the previous output as input, producing $\hat{\phi}_i^*$.
Training runs for 50 epochs with the Adam~\cite{adam} optimizer. The learning rate is linearly warmed up for the first 1\% of total steps and then kept constant at $10^{-5}$. Validation is performed after each epoch. We use a batch size of 64 for DistilGPT2 and 32 for the rest, with gradient accumulation. For models above 1B parameters, dropout with probability $\rho_{\mathrm{back}}=0.75$ is applied during backpropagation, this means that gradients are computed only for a random subset of documents per batch, while the rest use stop-gradient~\cite{tack2024mac}. LLaMA-2-7B is trained with 4-bit quantization~\cite{4bitquant} for both the model and the amortization network.
The codebook size is fixed to $N_c=512$, with VQ commitment cost $\beta_{\text{commit}}=0.25$ and weight $\lambda_{\mathrm{VQ}}=1.0$. Codebook resetting uses EMA decay rate $\gamma=0.99$ and reset threshold $\epsilon=10^{-4}$.
For KV-LoRA, DistilGPT2 uses $r=16$, $\alpha=32$, $\rho=0.05$, applied to the last $n_{\mathrm{lora}}=6$ layers. GPT2-Large, GPT2-XL and LLaMA-2-7B use $r=32$, $\alpha=64$, $\rho=0.05$, applied to the last $n_{\mathrm{lora}}=16$ layers. For the GPT-2 family, we share the LoRA down-projection matrix across $K$ and $V$, this means $A_{K,\ell}=A_{V,\ell}$.
All experiments are conducted on a single NVIDIA A100 80GB GPU.

\subsubsection{Examined Models}  
\begin{itemize}
    \item \textbf{Uniform Fine-Tuning}\footnote{\url{https://github.com/nathanhu0/CaMeLS}}: A baseline approach where all tokens in the new documents are treated equally during model updates.
    \item \textbf{Salient Spans}\footnotemark[1]~\cite{salientspansbaseline}: A heuristic-based method that fine-tunes only on tokens within pre-identified salient spans, ignoring the rest.
    \item \textbf{CaMeLS}\footnotemark[1]~\cite{huCamels}: Context-aware Meta-learned Loss Scaling, which uses a meta-trained network to assign importance weights to tokens during fine-tuning, focusing learning on the most informative content.
    \item \textbf{MAC}\footnote{\url{https://github.com/jihoontack/MAC}}~\cite{tack2024mac}: Memory of Amortized Contexts, an online adaptation framework that freezes the base model and uses a meta-learned network to encode documents into compact modulations stored in a memory bank. An aggregation module retrieves and combines the modulations with the query, without requiring gradient updates during inference.
    \item
    \textbf{MBC}\footnote{\url{https://github.com/Thomkat/MBC}}: The proposed model.
\end{itemize}

For fair comparison, we retrained all baselines. For Uniform Fine-Tuning, Salient Spans and CaMeLS, we followed the configuration of~\cite{huCamels}. Each pretrained LLM is first fine-tuned on QA pairs to obtain a task-adapted base model. During this pretraining, an inner batch of 6 document–query–label triples is used, and outer-loop gradients are accumulated over 24 examples, split into 4 batches of 6. Subsequently, the base model undergoes online adaptation, where it is updated on a stream of documents. The learning rate for each base–strategy combination is selected via a hyperparameter sweep on the validation set over $\{10^{-4}, 2.5\!\times\!10^{-5}, 6.25\!\times\!10^{-6}, 1.625\!\times\!10^{-6}\}$. The best learning rate for Uniform and Salient Spans is mostly $1.625\times 10^{-6}$, while for CaMeLS $2.5\times 10^{-5}$. Adam is used in most cases, and Adafactor~\cite{adafactor} is used for large models.
For MAC, we followed the configuration of~\cite{tack2024mac}. Specifically, the amortization network uses T5-Small (12 output tokens) for DistilGPT2, T5-Base (24 output tokens) for GPT2-Large, and T5-Large (24 output tokens) for GPT2-XL and LLaMA-2-7B, while the input encoder uses T5-Small for DistilGPT2 and T5-Base for the rest. The aggregation network consists of four cross-attention blocks. Training is performed for 50 epochs with Adam and a constant learning rate of $10^{-5}$ after a one-epoch warm-up, using a batch size of 64 for DistilGPT2 and 32 for the rest with gradient accumulation. For backbones exceeding 1B parameters, backpropagation dropout with probability $\rho_{\text{back}}=0.75$ is applied, and LLaMA-2-7B is trained with 4-bit quantization~\cite{4bitquant}.

\subsection{Experimental Results}

\begin{table*}[]
\centering
\small
\caption{Exact Match (EM) and F1 scores on StreamingQA, SQuAD, and ArchivalQA across different backbone models and baselines. High values indicate high QA performance.} 
\label{tab:mbc_results}
\renewcommand{\arraystretch}{1}
\begin{NiceTabular}{%
  @{}p{2.2cm}
  @{}l
  *{6}{@{\hspace{19pt}}c}
  @{}
}
\CodeBefore
  \cellcolor{rowgray}{7-2} \cellcolor{rowgray}{7-3} \cellcolor{rowgray}{7-4} \cellcolor{rowgray}{7-5} \cellcolor{rowgray}{7-6} \cellcolor{rowgray}{7-7} \cellcolor{rowgray}{7-8}
  \cellcolor{rowgray}{12-2} \cellcolor{rowgray}{12-3} \cellcolor{rowgray}{12-4} \cellcolor{rowgray}{12-5} \cellcolor{rowgray}{12-6} \cellcolor{rowgray}{12-7} \cellcolor{rowgray}{12-8}
  \cellcolor{rowgray}{17-2} \cellcolor{rowgray}{17-3} \cellcolor{rowgray}{17-4} \cellcolor{rowgray}{17-5} \cellcolor{rowgray}{17-6} \cellcolor{rowgray}{17-7} \cellcolor{rowgray}{17-8}
  \cellcolor{rowgray}{22-2} \cellcolor{rowgray}{22-3} \cellcolor{rowgray}{22-4} \cellcolor{rowgray}{22-5} \cellcolor{rowgray}{22-6} \cellcolor{rowgray}{22-7} \cellcolor{rowgray}{22-8}
\Body
\toprule
\multicolumn{1}{c}{\makecell{\textbf{Model} \\ \textbf{(\# params)}}} &
\multicolumn{1}{c}{\textbf{Method}} &
\multicolumn{2}{c}{\textbf{StreamingQA}} &
\multicolumn{2}{c}{\textbf{SQuAD}} &
\multicolumn{2}{c}{\textbf{ArchivalQA}} \\
\cmidrule(lr){3-4}\cmidrule(lr){5-6}\cmidrule(lr){7-8}
& & \textbf{EM (\(\uparrow\))} & \textbf{F1 (\(\uparrow\))} &
      \textbf{EM (\(\uparrow\))} & \textbf{F1 (\(\uparrow\))} &
      \textbf{EM (\(\uparrow\))} & \textbf{F1 (\(\uparrow\))} \\
\midrule

\Block{5-1}{\makecell[l]{DistilGPT2\\(82M)}}
  & Uniform        & 1.62 & 2.97  & 1.34 & 2.78 & 4.01 & 3.69 \\
  & Salient Spans  & 1.62 & 4.33  & 1.31 & 2.50 & 4.08 & 3.98 \\
  & CaMeLS         & 1.86 & 4.38  & 1.43 & 3.06 & 4.11 & 5.99 \\
  & MAC            & 3.48 & 8.11  & 1.90 & 5.00 & 5.99 & 8.87 \\
  & \textbf{MBC (Ours)}  & \makecell{\textbf{3.96}  \gain{3.96}{3.48}} & \makecell{\textbf{8.76}  \gain{8.76}{8.11}} & \makecell{\textbf{2.10}  \gain{2.10}{1.90}} & \makecell{\textbf{5.36}  \gain{5.36}{5.00}} & \makecell{\textbf{6.61}  \gain{6.61}{5.99}} & \makecell{\textbf{9.27}  \gain{9.27}{8.87}} \\
\midrule

\Block{5-1}{\makecell[l]{GPT2-Large\\(774M)}}
  & Uniform        & 4.14 & 8.08  & 3.37 & 5.62 & 8.03 & 6.63 \\
  & Salient Spans  & 4.26 & 8.53  & 4.38 & 6.79 & 9.75 & 7.23 \\
  & CaMeLS         & 5.48 & 10.31 & 4.45 & 7.60 & 9.28 & 9.18 \\
  & MAC            & 6.12 & 11.44 & 6.14 & 9.75 & 10.95 & 12.15 \\
  & \textbf{MBC (Ours)}  & \makecell{\textbf{7.43}  \gain{7.43}{6.12}}& \makecell{\textbf{12.77}  \gain{12.77}{11.44}} & \makecell{\textbf{6.99}  \gain{6.99}{6.14}} & \makecell{\textbf{10.88}  \gain{10.88}{9.75}} & \makecell{\textbf{12.03}  \gain{12.03}{10.95}} & \makecell{\textbf{13.68}  \gain{13.68}{12.15}} \\
\midrule

\Block{5-1}{\makecell[l]{GPT2-XL\\(1.5B)}}
  & Uniform        & 5.16 & 9.14  & 5.87 & 7.87 & 9.89 & 10.46 \\
  & Salient Spans  & 5.46 & 11.32  & 5.66 & 8.69 & 10.44 & 13.68 \\
  & CaMeLS         & 6.98 & 11.23 & 6.17 & 9.93 & 11.48 & 14.01 \\
  & MAC            & 7.14 & 12.01 & 6.89 & 10.12 & 11.48 & 15.52 \\
  & \textbf{MBC (Ours)}  & \makecell{\textbf{7.49}  \gain{7.49}{7.14}} & \makecell{\textbf{12.77}  \gain{12.77}{12.01}} & \makecell{\textbf{7.40}  \gain{7.40}{6.89}} & \makecell{\textbf{11.96}  \gain{11.96}{10.12}} & \makecell{\textbf{12.34}  \gain{12.34}{11.48}} & \makecell{\textbf{15.93}  \gain{15.93}{15.52}} \\
\midrule

\Block{5-1}{\makecell[l]{LLaMA-2\\(7B)}}
  & Uniform        & 11.76 & 12.53 & 12.78 & 16.62 & 17.89 & 20.01 \\
  & Salient Spans  & 12.12 & 18.65 & 13.32 & 18.09 & 18.45 & 22.21 \\
  & CaMeLS*         & N/A & N/A & N/A & N/A & N/A & N/A \\
  & MAC            & 14.01 & 20.44 & 13.33 & 18.17 & 19.58 & 23.89 \\
  & \textbf{MBC (Ours)}  & \makecell{\textbf{16.04}  \gain{16.04}{14.01}} & \makecell{\textbf{25.33}  \gain{25.33}{20.44}} & \makecell{\textbf{14.93}  \gain{14.93}{13.33}} & \makecell{\textbf{22.15}  \gain{22.15}{18.17}} & \makecell{\textbf{22.71}  \gain{22.71}{19.58}} & \makecell{\textbf{28.66}  \gain{28.66}{23.90}} \\
\bottomrule
\end{NiceTabular}
\\[0.5em]
\parbox{\textwidth}{\footnotesize
\textit{* CaMeLS results are not reported for LLaMA-2 (7B) because the model exceeds the memory capacity of a single NVIDIA A100 80GB GPU. Even with a batch size of 1, it was infeasible to replicate this baseline under our hardware constraints.}
}
\end{table*}

\begin{figure*}[]
  \centering
  \includegraphics[width=\linewidth]{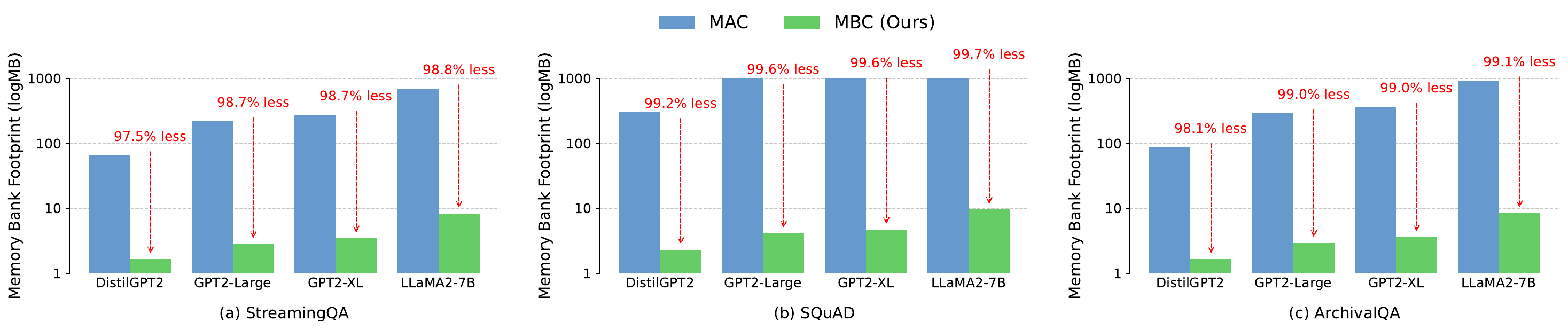}
  \caption{Memory bank footprint (logMB) of MAC and MBC across StreamingQA, SQuAD, and ArchivalQA.}
  \label{fig:memorysavings}
\end{figure*}

\subsubsection{QA Performance Evaluation}
Table~\ref{tab:mbc_results} reports the QA performance. We compare MBC against the baseline methodologies, with MAC being the most competitive. Across all datasets and base LLMs, MBC consistently improves both EM and F1. On average, MBC gains 11.84\% in EM and 12.99\% in F1 compared to MAC. The performance gains result from two main design choices. Firstly, the introduction of KV-LoRA allows the attention mechanism to make  effective use of the modulation $\hat{\phi}_i^*$, leading to  accurate answers. In addition, an efficiently learned codebook preserves the quality of the stored documents, ensuring that the compression mechanism does not degrade the performance.

\begin{table}[t]
\centering
\caption{Trainable parameters of MAC and MBC (offline).}
\label{tab:trainable_params}
\resizebox{\columnwidth}{!}{%
\begin{tabular}{@{}lcccc@{}}
\toprule
 Method & DistilGPT2 & GPT2-Large & GPT2-XL & LLaMA-2-7B \\
\midrule
MAC & 197M & 927M & 1.72B & 2.36B \\
MBC & 197.6M (+0.31\%) & 929.6M (+0.28\%) & 1.723B (+0.19\%) & 2.371B (+0.45\%) \\
\bottomrule
\end{tabular}%
}
\end{table}

\begin{table*}[] 
\centering 
\setlength{\tabcolsep}{6pt} 
\renewcommand{\arraystretch}{0.95} 
\caption{Memory bank size (MB) / F1 retention rate (\%) on StreamingQA, SQuAD, and ArchivalQA for different base LLMs. Each entry is reported as memory bank footprint in MB followed by the corresponding retention rate.} 

\resizebox{\textwidth}{!}{% 

\begin{tabular}{@{}l*{6}{c}|*{6}{c}@{}} 
\hline 
\multicolumn{7}{c|}{\textbf{DistilGPT2}} & \multicolumn{6}{c}{\textbf{GPT2-Large}}\\ 
\cline{1-7}\cline{8-13} 
\multicolumn{1}{c}{\textbf{\# of Doc}} & 
\multicolumn{2}{c}{\textbf{StreamingQA}} & 
\multicolumn{2}{c}{\textbf{SQuAD}} & 
\multicolumn{2}{c|}{\textbf{ArchivalQA}} & 
\multicolumn{2}{c}{\textbf{StreamingQA}} & 
\multicolumn{2}{c}{\textbf{SQuAD}} & 
\multicolumn{2}{c}{\textbf{ArchivalQA}}\\ 
\cline{2-7}\cline{8-13} 
& \textbf{MAC} & \textbf{MBC} & \textbf{MAC} & \textbf{MBC} & \textbf{MAC} & 
\multicolumn{1}{c|}{\textbf{MBC}} & 
\textbf{MAC} & \textbf{MBC} & \textbf{MAC} & \textbf{MBC} & \textbf{MAC} & \textbf{MBC}\\ 
\hline 
200  & 8.21/100 & 1.52/100 & 38.1/100 & 1.60/100 & 10.91/100 & 1.53/100 & 27.36/100 & 2.54/100 & 126.99/100 & 2.7/100 & 36.36/100 & 2.56/100 \\ 
400  & 16.42/99.5 & 1.54/99 & 76.19/99.5 & 1.70/99 & 21.82/99.5 & 1.56/99 & 54.73/99 & 2.59/99 & 253.98/99 & 2.9/99.5 & 72.72/99 & 2.61/99 \\ 
600  & 24.63/99 & 1.56/99 & 114.29/99 & 1.80/99 & 32.72/99.5 & 1.59/99.5 & 82.09/99 & 2.63/99 & 380.96/98.5 & 3.1/99.5 & 109.07/99 & 2.67/99 \\ 
800  & 32.84/99 & 1.59/99 & 152.39/99 & 1.90/98 & 43.63/99 & 1.62/99 & 109.46/98 & 2.67/99 & 507.95/98 & 3.3/99 & 145.43/98.8 & 2.73/99 \\ 
1000 & 41.04/98 & 1.61/98 & 190.48/99 & 1.99/97.5 & 54.54/99 & 1.64/98.5 & 136.82/97.5 & 2.71/98.5 & 634.94/97.5 & 3.49/98.5 & 181.79/97.5 & 2.78/98.8 \\ 
1200 & 49.25/98 & 1.63/98 & 228.58/98.5 & 2.09/98 & 65.45/98.5 & 1.67/98 & 164.18/97 & 2.76/98.5 & 761.93/97 & 3.69/98 & 218.15/97.5 & 2.84/98.3 \\ 
1400 & 57.46/98 & 1.65/98.5 & 266.67/98 & 2.19/98.5 & 76.35/98.5 & 1.70/98.5 & 191.55/96.5 & 2.8/97 & 888.91/96.5 & 3.89/96.5 & 254.5/96.5 & 2.89/97.6 \\ 
1600 & 65.67/98 & 1.67/98 & 304.77/98 & 2.29/98.5 & 87.26/98 & 1.73/98 & 218.91/97 & 2.84/97 & 1015.9/96 & 4.09/97 & 290.86/97 & 2.95/97.2 \\ 
\hline 
\end{tabular}}% 

\vspace{4pt}

\resizebox{\textwidth}{!}{% 
\begin{tabular}{@{}l*{6}{c}|*{6}{c}@{}} 
\hline 
\multicolumn{7}{c|}{\textbf{GPT2-XL}} & \multicolumn{6}{c}{\textbf{Llama2-7B}}\\ 
\cline{1-7}\cline{8-13} 
\multicolumn{1}{c}{\textbf{\# of Doc}} & 
\multicolumn{2}{c}{\textbf{StreamingQA}} & 
\multicolumn{2}{c}{\textbf{SQuAD}} & 
\multicolumn{2}{c|}{\textbf{ArchivalQA}} & 
\multicolumn{2}{c}{\textbf{StreamingQA}} & 
\multicolumn{2}{c}{\textbf{SQuAD}} & 
\multicolumn{2}{c}{\textbf{ArchivalQA}}\\ 
\cline{2-7}\cline{8-13} 
& \textbf{MAC} & \textbf{MBC} & \textbf{MAC} & \textbf{MBC} & \textbf{MAC} & 
\multicolumn{1}{c|}{\textbf{MBC}} & 
\textbf{MAC} & \textbf{MBC} & \textbf{MAC} & \textbf{MBC} & \textbf{MAC} & \textbf{MBC}\\ 
\hline 
200  & 34.2/100 & 3.16/100 & 158.73/100 & 3.32/100 & 45.45/100 & 3.18/100 & 87.56/100 & 8.04/100 & 406.36/100 & 8.2/100 & 116.34/100 & 8.06/100 \\ 
400  & 68.41/100 & 3.21/100 & 317.47/99.5 & 3.52/99 & 90.89/100 & 3.23/99 & 175.13/100 & 8.09/99 & 812.72/99.7 & 8.4/99.5 & 232.69/99 & 8.11/99.5 \\ 
600  & 102.61/99 & 3.25/98.5 & 476.2/98.8 & 3.72/99 & 136.34/99.7 & 3.29/99 & 262.69/99.5 & 8.13/99 & 1219.08/99.5 & 8.6/99 & 349.03/99.8 & 8.17/99 \\ 
800  & 136.82/98 & 3.29/99 & 634.94/98.0 & 3.92/98.7 & 181.79/99.4 & 3.35/98.6 & 350.25/98 & 8.17/98.5 & 1625.44/99.5 & 8.8/98.5 & 465.38/99.5 & 8.23/98.5 \\ 
1000 & 171.02/97 & 3.33/98.5 & 793.67/98.0 & 4.11/98.8 & 227.23/98.5 & 3.4/98.1 & 437.81/97.4 & 8.21/98 & 2031.8/99 & 8.99/97.5 & 581.72/98.5 & 8.28/98.2 \\ 
1200 & 205.22/96.7 & 3.38/98 & 952.4/97.9 & 4.31/98 & 272.68/98 & 3.46/97.5 & 525.38/97 & 8.26/97 & 2438.16/97 & 9.19/96.9 & 698.06/98.2 & 8.34/98 \\ 
1400 & 239.43/96 & 3.42/97 & 1111.14/98.0 & 4.51/98 & 318.12/98.3 & 3.51/97.8 & 612.94/96.5 & 8.3/96.5 & 2844.52/96.7 & 9.39/96.3 & 814.41/98 & 8.39/98 \\ 
1600 & 273.63/95.5 & 3.46/96 & 1269.87/97.5 & 4.71/97.5 & 363.57/98 & 3.58/97.5 & 700.5/95 & 8.34/96 & 3250.88/96.5 & 9.59/96.5 & 930.75/97.5 & 8.45/97.5 \\ 
\hline 
\end{tabular}}% 
\label{tab:retention} 
\end{table*}

\subsubsection{Memory Bank Size}
Figure~\ref{fig:memorysavings} compares the memory bank footprint of two examined memory-augmented methods, that is MBC and MAC.
For MBC, the footprint includes the codebook and stored indices in the bank, while for MAC it corresponds to the full memory bank.
Across all three datasets, MBC achieves substantial memory savings. For DistilGPT2, the memory bank size is reduced by an average of 98.27\% compared to MAC. For GPT2-Large and GPT2-XL, the reduction averages 99.1\%, and for LLaMA-2-7B it averages 99.2\%. These results show that memory compression is consistently effective across all the different model scales.

To examine the overhead of the codebook and KV-LoRA in MBC, we compare the trainable parameters of the two examined memory augmentation methods, namely MAC and MBC. These numbers are reported in the offline setting, this means without considering documents stored during online adaptation.
As Table~\ref{tab:trainable_params} shows, the additional parameters introduced by the codebook and KV-LoRA of the proposed MBC model account for less than 0.5\% across all base LLMs. 
This overhead is negligible compared to the improvements in the QA accuracy and memory compression.

\subsubsection{Knowledge Retention During Online Adaptation}  
In this experiment, we evaluate how well models retain knowledge from previously adapted documents while continuing to adapt to new documents. Following the evaluation protocol in online adaptation from~\cite{tack2024mac}, we measure the F1 score retention rate, defined as the relative decline in the F1 score of the first 200 adapted documents after further adaptation on up to 1,400 additional documents, with a step of 200 documents. A high retention rate indicates that the model preserves knowledge from earlier documents even as new information is incorporated, showing reduced susceptibility to catastrophic forgetting.
Table~\ref{tab:retention} reports the retention rates alongside the corresponding memory bank sizes. MBC achieves high retention, comparable to MAC, across all base models and datasets, demonstrating that compression does not harm the model’s ability to preserve earlier knowledge during online adaptation. At the same time, MBC consistently requires far less memory. For the same number of adapted documents, its memory bank footprint is reduced on average by 97.3\% compared to MAC.
Interestingly, the small model DistilGPT2 appears to show high retention. However, this effect comes from its limited capacity to utilize the memory bank effectively: its absolute performance is already low, so retention appears artificially high. Meanwhile, the large models GPT2-XL and LLaMA-2-7B achieve strong adaptation performance and high retention, confirming that MBC scales effectively to large base LLMs.

\begin{figure}[t]
  \centering
  \includegraphics[width=0.96\linewidth]{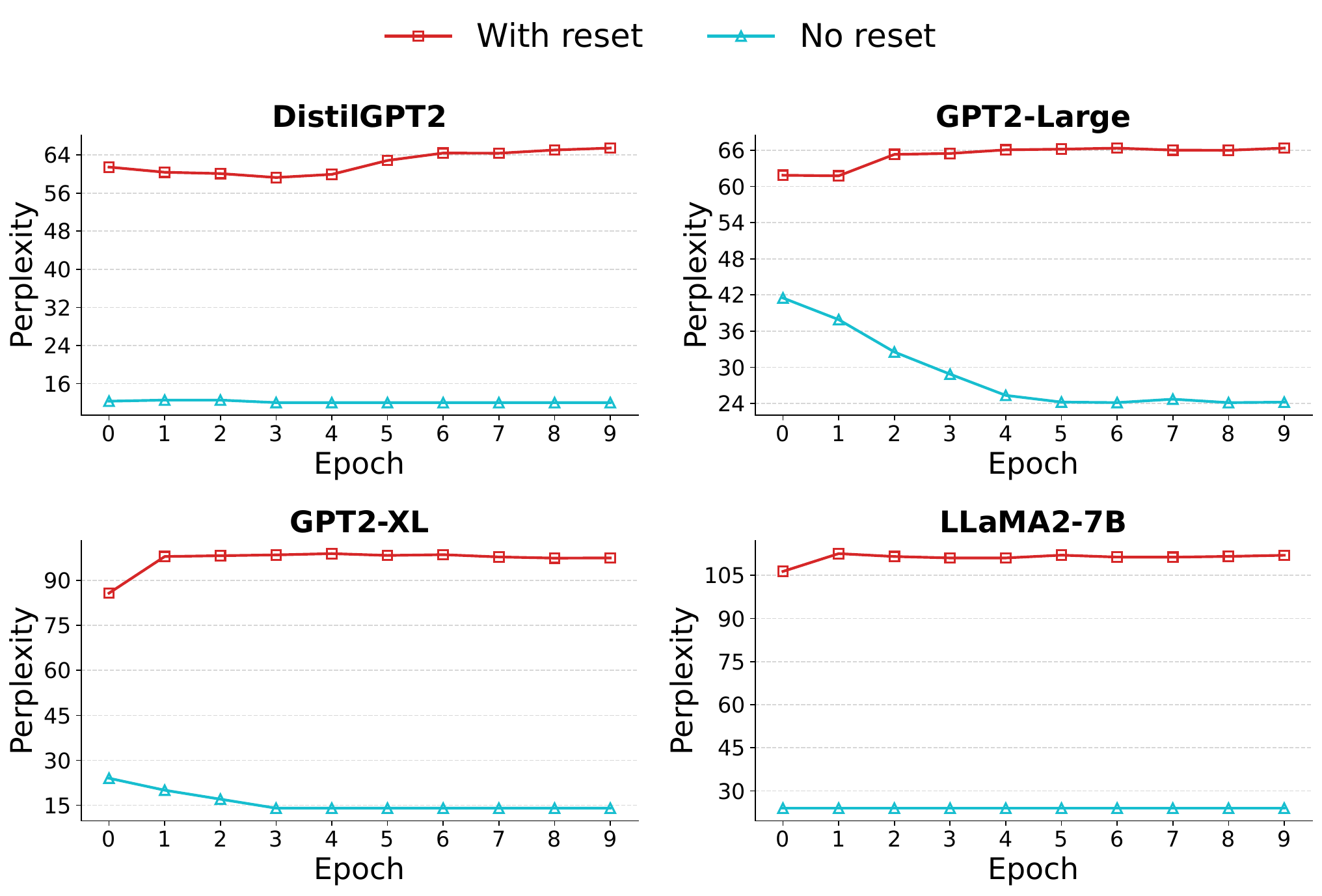}
  \caption{Perplexity over train epochs on StreamingQA with and without codebook resetting in MBC, across all base LLMs.}
  \label{fig:perplexity}
\end{figure}

\subsubsection{Effectiveness of the Codebook Resetting Mechanism}
\label{sec:codebook_reset}
We further evaluate the role of the EMA-based codebook resetting mechanism introduced in Section~\ref{sec:codebook_reset_intro} by comparing training runs with and without resetting in MBC. Code usage is measured via perplexity, defined as \(\mathrm{PPL} = \exp\!\big(-\sum_k \bar p_k \log \bar p_k\big)\), where \(\bar p_k = u_k / \sum_j u_j\) and \(u_j\) is the EMA-smoothed usage defined in Eq.~\ref{eq:resetting}. Lower values that remain flat indicate codebook collapse, this means that only a small subset of codes being repeatedly used.
Figure~\ref{fig:perplexity} shows the perplexity curves on StreamingQA across the four base LLMs. With resetting, effective code usage remains stable and diverse throughout training. For DistilGPT2, perplexity is between 57 and 65 during the first 10 epochs, while without resetting it collapses close to 12. For GPT2-Large, resetting maintains perplexity between 61–66, whereas without resetting it quickly drops to 24. For GPT2-XL, resetting maintains perplexity steadily above 90, whereas it collapses to 14 without resetting. Similarly, for LLaMA-2-7B, resetting maintains perplexity above 100, while without it the codebook again collapses to 24.
These results confirm that the codebook resetting mechanism is important for preventing collapse and ensuring balanced code usage, which supports stable training and effective adaptation for the proposed MBC method.

\section{Conclusion}
\label{sec:Conclusion}
In this work, we addressed the scalability challenges of memory-augmented LLMs, where the memory bank grows constantly as new documents are processed. We proposed MBC, a model that compresses the memory bank, enabling efficient continual adaptation of LLMs in streaming settings. By combining codebook-based compression with an online resetting mechanism, MBC prevents codebook collapse and ensures balanced code utilization. At the same time, lightweight KV-LoRA modules provide targeted adaptation within the attention mechanism, allowing the model to efficiently exploit the query-memory modulations without full fine-tuning. This design enables MBC to achieve scalability in terms of memory efficiency while improving the QA accuracy. Experiments with QA datasets demonstrate that MBC improves EM and F1 score while reducing the memory bank footprint to 0.3\% of the most competitive baseline. MBC also maintains high F1 retention during online adaptation, thus reducing catastrophic forgetting. An interesting future direction is to extend MBC by incorporating reinforcement signals to guide memory usage adaptively~\cite{kulkarniReinforcementLearningOptimizing2024} or by exploring distributed memory banks that enable federated continual learning~\cite{yang2024federated}.


\printbibliography

\end{document}